\documentclass[conference]{IEEEtran}
\IEEEoverridecommandlockouts

\usepackage{cite}
\usepackage{amsmath,amssymb,amsfonts}
\usepackage{algorithmic}
\usepackage{graphicx}
\usepackage{textcomp}
\usepackage{xcolor}

\usepackage{booktabs}
\usepackage{multirow}
\usepackage{flushend}
\usepackage{stfloats}
\usepackage[colorlinks,
linkcolor=red,
anchorcolor=blue,
citecolor=green]{hyperref}
\usepackage{tikz}
\usetikzlibrary{shapes,arrows}
\def\BibTeX{{\rm B\kern-.05em{\sc i\kern-.025em b}\kern-.08em
T\kern-.1667em\lower.7ex\hbox{E}\kern-.125emX}}

\begin{document}

\title{Enhancing Chest X-ray Classification through Knowledge Injection in Cross-Modality Learning\\
	\thanks{This study has received ethics approval from the Research Ethics Committee of SAHZU (IRB-2023-1109), and funded by the Westlake University Research Center for Industries of the Future (Grant No.WU2023C017). We are also grateful to the Google Developer Relations Team for offering GCP credits.}
}

\author{
	\IEEEauthorblockN{
		Yang Yan\textsuperscript{\tiny $\blacklozenge\clubsuit\diamondsuit\heartsuit$},
		Bingqing Yue\textsuperscript{\tiny $\blacklozenge\bigstar\heartsuit$},
		Qiaxuan Li\textsuperscript{\tiny $\blacklozenge\bigstar$},
		Man Huang\textsuperscript{\tiny $\blacklozenge\bigstar$},
		Jingyu Chen\textsuperscript{\tiny $\blacklozenge\bigstar$},
		and Zhenzhong Lan\textsuperscript{\tiny $\clubsuit\diamondsuit$}\thanks{Corresponding to: lanzhenzhong@westlake.edu.cn}
	}
	\IEEEauthorblockA{\textsuperscript{\tiny $\blacklozenge$} Zhejiang University, Hangzhou, Zhejiang, China}
	\IEEEauthorblockA{\textsuperscript{\tiny $\clubsuit$} School of Engineering, Westlake University, Hangzhou, Zhejiang, China}
	\IEEEauthorblockA{\textsuperscript{\tiny $\diamondsuit$} Westlake University Research Center for Industries of the Future, Westlake University, Hangzhou, Zhejiang, China}
	\IEEEauthorblockA{\textsuperscript{\tiny $\bigstar$} The Second Affiliated Hospital Zhejiang University School of Medicine (SAHZU), Hangzhou, Zhejiang, China}
	\IEEEauthorblockA{\{yan.yang, lucyybq, liqiaxuan, huangman, chenjy9611\}@zju.edu.cn, \{yanyang, lanzhenzhong\}@westlake.edu.cn}
}

\maketitle
\begingroup\renewcommand\thefootnote{$\heartsuit$}

\footnotetext{Equal contribution}
\endgroup

\begin{abstract}
	The integration of artificial intelligence in medical imaging has shown tremendous potential, yet the relationship between pre-trained knowledge and performance in cross-modality learning remains unclear. This study investigates how explicitly injecting medical knowledge into the learning process affects the performance of cross-modality classification, focusing on Chest X-ray (CXR) images. We introduce a novel Set Theory-based knowledge injection framework that generates captions for CXR images with controllable knowledge granularity. Using this framework, we fine-tune CLIP model on captions with varying levels of medical information. We evaluate the model's performance through zero-shot classification on the CheXpert dataset, a benchmark for CXR classification. Our results demonstrate that injecting fine-grained medical knowledge substantially improves classification accuracy, achieving 72.5\% compared to 49.9\% when using human-generated captions. This highlights the crucial role of domain-specific knowledge in medical cross-modality learning. Furthermore, we explore the influence of knowledge density and the use of domain-specific Large Language Models (LLMs) for caption generation, finding that denser knowledge and specialized LLMs contribute to enhanced performance. This research advances medical image analysis by demonstrating the effectiveness of knowledge injection for improving automated CXR classification, paving the way for more accurate and reliable diagnostic tools.
\end{abstract}


\section{Introduction}

The advent of pre-trained models has revolutionized both Natural Language Processing and Computer Vision~\cite{jia2021align,radford2021clip,yang2022tcl,NEURIPS2022_960a172b}, demonstrating the power of leveraging extensive knowledge for enhanced performance on downstream tasks. While this paradigm has shown promising results in various domains, its application in medical cross-modality learning, particularly in critical areas like medical imaging, remains unexplored.

Chest X-ray (CXR) image classification offers a compelling testbed for investigating the impact of knowledge injection in cross-modality learning. CXR images are standardized~\cite{jones2015chest}, and common diseases have well-defined textual descriptions~\cite{hansell2008fleischner}, providing a robust framework for analyzing the relationship between knowledge and model performance. Current approaches often rely on human-generated captions for training, which may lack the granularity and precision required to fully capture the rich medical knowledge embedded in these images. This raises a crucial question: \textbf{Can explicitly injecting medical knowledge into the learning process significantly enhance the performance of cross-modality classification for CXR images?}

\begin{figure}[t]
	\centering
	\resizebox{0.6\linewidth}{!}{
		\begin{tikzpicture}[node distance=1.5cm, auto]
			\tikzstyle{block} = [rectangle, draw, fill=white, 
			text width=4em, text centered, rounded corners, minimum height=2em, font=\small]
			\tikzstyle{line} = [draw, -latex']
			
			\node [block] (labels) {Disease Labels};
			\node [block, below of=labels] (image) {CXR Image};
			\node [block, right of=image, xshift=3cm] (caption) {Caption Text};
			
			\path [line] (image) -- (labels);
			\path [line] (image) -- (caption);
			\path [line] (labels) -- (caption);
			
			\node [above of=caption, xshift=-1.1cm, text width=12em, text centered, font=\footnotesize] 
			{\textbf{Proposed Method}\\+ Injected Knowledge with prior and domain-specific LLM\\(Rich Knowledge)};
			\node [below of=caption, xshift=-2.3cm, yshift=0.8cm, text width=10em, text centered, color=red!80!black, font=\footnotesize] 
			{\textbf{Traditional Method}\\Human/vLLM caption (limited knowledge)};
			
		\end{tikzpicture}
	}
	\caption{Illustration of knowledge injection in cross-modality learning.}
	\label{fig:knowledge_injection}
\end{figure}
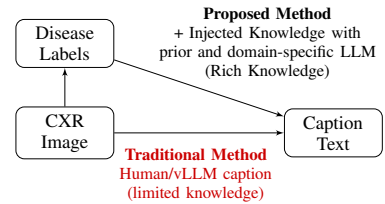

This research directly addresses this question by investigating the impact of knowledge injection on CXR classification. We hypothesize that incorporating medical knowledge with controllable granularity will empower models to better understand and interpret medical images, leading to improved classification accuracy. To systematically explore this hypothesis, we investigate the following research questions:
1. Are human-generated captions sufficiently knowledgeable for effective medical cross-modality learning?
2. How can we inject knowledge into cross-modality learning models?
3. What factors influence the effectiveness of knowledge injection in this context?

To address these questions, we propose a novel Set Theory-based framework for injecting medical knowledge into CXR captions with controllable granularity, ranging from basic disease labels to detailed descriptions of phenotypes. We fine-tune a pre-trained CLIP model using these generated captions and evaluate its zero-shot classification performance on the CheXpert dataset. This work contributes: (1) a novel framework for knowledge injection with controllable granularity, (2) evidence that fine-grained knowledge injection significantly improves CXR classification accuracy, and (3) insights into the influence of knowledge density and the choice of LLMs for caption generation.

\section{Methodology}

To investigate the impact of knowledge injection on medical cross-modality learning, we focus on CXR image classification. This choice is motivated by the standardized nature of CXR acquisition~\cite{jones2015chest} and the well-defined textual descriptions of common lung diseases~\cite{hansell2008fleischner}, which provide a suitable testbed for evaluating our hypothesis. We leverage two publicly available datasets: MIMIC-CXR~\cite{johnson2019mimic,johnson2019mimiccxrjpg} and CheXpert~\cite{irvin2019chexpert}. The MIMIC-CXR dataset, with its associated radiological reports, serves as the source for generating captions with varying levels of injected knowledge. The CheXpert dataset, containing CXR images and expert-labeled observations for 14 common chest radiographic observations, is used for evaluating the performance of our fine-tuned CLIP model through zero-shot classification. We utilize accuracy as our primary evaluation metric, calculated as the average per-class accuracy across 11 common lung diseases.

Our methodology involves a two-step process: (1) developing a Set Theory-based knowledge injection framework to generate captions with controllable knowledge granularity and (2) fine-tuning a pre-trained CLIP model~\cite{radford2021clip} using these generated captions for cross-modality classification.

\subsection{Set Theory-based Knowledge Injection Framework}
\label{sec:sec:knowledge_injection_framework}

We propose a novel framework grounded in Set Theory to inject medical knowledge into CXR image captions with varying levels of detail. This framework leverages expert knowledge to define relationships between diseases ($D$) and their associated phenotypes ($P$). For each disease $d_i \in D$, we define a set of typical phenotypes $P_i^{typ}$ and a set of excluded phenotypes $P_i^{exc}$. These sets are then used to guide the generation of captions with three levels of granularity:

\begin{enumerate}
	\item \textbf{Coarse-grained:} Only the disease label $d_i$ is included (e.g., "Pneumonia").
	\item \textbf{Medium-grained:} The caption includes $d_i$ and its typical phenotypes $P_i^{typ}$ (e.g., "Pneumonia with consolidation in the lower lobe").
	\item \textbf{Fine-grained:} The caption includes $d_i$, $P_i^{typ}$, and explicitly mentions the absence of phenotypes in $P_i^{exc}$ (e.g., "Pneumonia with consolidation in the lower lobe. No evidence of pneumothorax or pleural effusion").
\end{enumerate}

We utilize different LLMs to paraphrase the information into natural language captions, ensuring diversity and reducing potential biases. This framework enables us to systematically investigate the impact of knowledge density on the performance of the fine-tuned CLIP model.

\subsection{Cross-modality Classification with CLIP}

We employ CLIP model, a powerful dual-encoder architecture pre-trained on a massive dataset of image-text pairs, for our cross-modality classification task. CLIP's ability to learn robust image-text alignments and its proficiency in zero-shot learning make it particularly well-suited for evaluating the effects of our knowledge injection strategy.

We fine-tune CLIP using the generated captions with different knowledge granularities. The fine-tuning process adapts the pre-trained CLIP model to the specific task of CXR image classification by further training it on our dataset of CXR images and their corresponding generated captions. The training objective is to minimize the contrastive loss function (Eq.~\ref{eq:clip_loss}), which encourages the model to learn similar embeddings for matching image-text pairs and dissimilar embeddings for non-matching pairs.

\begin{equation}
	\resizebox{.91\hsize}{!}{$
		\mathcal{L}_{\text{CLIP}} = \frac{1}{N} \sum_{i=1}^N \left[ -\log \frac{\exp(I_i^T T_i / \tau)}{\sum_{j=1}^N \exp(I_i^T T_j / \tau)} - \log \frac{\exp(I_i^T T_i / \tau)}{\sum_{j=1}^N \exp(I_j^T T_i / \tau)} \right]
	$}
	\label{eq:clip_loss}
\end{equation}

After fine-tuning, we evaluate the model's performance using zero-shot classification. This involves encoding a given image into an embedding and comparing it to text embeddings representing the presence and absence of each disease label. The prediction for each label is determined by the highest cosine similarity between the image embedding and the corresponding text embeddings (Eq.~\ref{ref:zero_shot_classification}).

\begin{equation}
	\text{prediction} = \arg\max_{i \in \{1, ..., k\}} \text{cosine\_similarity}(I, t_i)
	\label{ref:zero_shot_classification}
\end{equation}

As CLIP's pre-training objective aligns with zero-shot learning, we can evaluate the impact of knowledge injection on cross-modality learning by measuring CLIP's performance on zero-shot tasks. We assess the model's performance through zero-shot classification, quantifying the effectiveness of our knowledge injection strategy.

\section{Experiments}

\subsection{Data Source and Evaluation Metric}

We utilize two publicly available CXR datasets: MIMIC-CXR~\cite{johnson2019mimiccxrjpg}, containing 377,110 images with radiologist annotations, and CheXpert~\cite{irvin2019chexpert}, comprising 224,316 images with disease labels extracted from reports. MIMIC-CXR serves as the source for generating captions with injected knowledge, while CheXpert is used for evaluating zero-shot classification accuracy, averaged across 11 common lung-related disease.

\subsection{Results}

\subsubsection{RQ1: Existing Knowledge is Insufficient for Effective Medical Cross-Modality Learning}

\begin{table}[t]
	\centering
	\label{tab:knowledge_source_comparison}
	\caption{{\small The average accuracy of zero-shot classification for 11 lung-related disease classes from CheXpert is presented.}}
	\resizebox{\linewidth}{!}{
		\begin{tabular}{llr}
			\toprule
			Model                                         & Knowledge Source                                                          & Acc. \\
			\midrule
			CLIP~\cite{radford2021clip}                   & General Pre-training Data                                                 & 43.9 \\
			\midrule
			Human Expert                                  & \em + Radiologist Annotation from MIMIC-CXR~\cite{johnson2019mimiccxrjpg} & 49.9 \\
			PubmedCLIP~\cite{eslami-etal-2023-pubmedclip} & \em + Abstract of Medical Papers                                          & 55.4 \\
			\midrule
			Ours                                          & \em + Injected Knowledge, Paraphrased by PaLM2~\cite{anil2023palm}        & 72.5 \\
			\bottomrule
		\end{tabular}
	}
\end{table}

To assess the sufficiency of existing knowledge for cross-modality learning in medical imaging, we evaluated the zero-shot classification performance of various pre-trained CLIP models on the CheXpert dataset. This dataset, with its standardized nature and well-defined disease labels, serves as a robust benchmark for examining the impact of pre-trained knowledge.

We began with the vanilla CLIP model~\cite{radford2021clip}, pre-trained on a massive, general image-text dataset, representing a model without exposure to specific medical knowledge. We then compared its performance to PubmedCLIP~\cite{eslami-etal-2023-pubmedclip}, a CLIP variant further pre-trained on PubMed abstracts, representing a model exposed to general medical knowledge. 

Finally, to assess the impact of CXR-specific knowledge, we fine-tuned vanilla CLIP on two caption sources from the MIMIC-CXR dataset~\cite{johnson2019mimiccxrjpg}: 1) human-generated captions, reflecting a common practice in medical cross-modality learning, and 2) captions generated using our proposed Set Theory-based knowledge injection framework, offering controlled injection of medical knowledge with fine-grained detail. 

As shown in Tab.~\ref{tab:knowledge_source_comparison}, the vanilla CLIP model performs poorly (43.9\% acc), highlighting the need for domain-specific knowledge. PubmedCLIP, benefiting from exposure to general medical knowledge, achieves higher accuracy (55.4\%). However, both are significantly outperformed by CLIP fine-tuned on our generated captions with fine-grained knowledge injection (72.5\% acc). This underscores that explicitly injecting medical knowledge into the pre-training process significantly enhances cross-modality classification performance, supporting our core hypothesis. Notably, even CLIP fine-tuned on human captions (49.9\% acc) underperforms PubmedCLIP, suggesting that human-generated captions may lack the detailed, knowledge necessary for optimal performance in this context, as Radiologist always take the normal phenotypes for granted and do not explicitly mention them in the captions.

\begin{table*}[t]
	\centering
	\caption{{\small Zero-shot classification accuracy on the CheXpert dataset for different levels of injected knowledge density. A: Atelectasis, CM: Cardiomegaly, C: Consolidation, E: Edema, ECM: Enlarged Cardiomediastinum, LL: Lung Lesion, LO: Lung Opacity, PE: Pleural Effusion, PO: Pleural Other, P: Pneumonia, PTX: Pneumothorax. }} 
	\label{tab:differrnt_knowledge_density}
	\resizebox{0.7\linewidth}{!}{
		\begin{tabular}{llrrrrrrrrrrrr}
			\toprule
			                                                        & diagnosis & A    & CM   & C    & E    & ECM  & LL   & LO   & PE   & PO   & P      & PTX  & Avg  \\
			Model                                                   & Grained   &      &      &      &      &      &      &      &      &      &        &      &      \\
			\midrule
			CLIP~\cite{radford2021clip}                             & -         & 59.9 & 32.8 & 36.2 & 38.4 & 51.7 & 62.1 & 63.4 & 68.5 & 61.6 & 5.2    & 3.4  & 43.9 \\
			PubmedCLIP~\cite{eslami-etal-2023-pubmedclip}           & -         & 65.9 & 32.3 & 53.9 & 19.8 & 52.6 & 99.6 & 53.9 & 44.0 & 98.3 & 79.3   & 9.5  & 55.4 \\
			\midrule
			Human Expert~\cite{johnson2019mimiccxrjpg}              & -         & 34.5 & 29.3 & 64.2 & 54.7 & 54.3 & 56.0 & 66.8 & 65.9 & 44.8 & 39.7   & 39.2 & 49.9 \\
			Palm2-bison~\cite{anil2023palm}                         & fine      & 74.6 & 70.7 & 72.4 & 77.6 & 56.9 & 84.5 & 70.3 & 80.2 & 93.1 & 37.1   & 79.7 & 72.5 \\
			\midrule
			\multirow[t]{3}{*}{llama2-7b}~\cite{touvron2023llama2}  & fine      & 47.8 & 41.4 & 72.4 & 44.0 & 45.3 & 84.9 & 44.4 & 69.8 & 79.3 & 42.7   & 70.7 & 58.4 \\
			                                                        & medium    & 69.4 & 55.2 & 41.8 & 58.2 & 61.2 & 66.8 & 68.5 & 63.4 & 83.6 & 3.4    & 12.1 & 53.1 \\
			                                                        & coarse    & 65.9 & 29.3 & 79.3 & 19.4 & 45.3 & 90.5 & 48.7 & 71.6 & 84.9 & 3.4    & 3.4  & 49.2 \\
			\cline{2-14}
			\multirow[t]{3}{*}{vicuna-7b}~\cite{zheng2023judging}   & fine      & 66.8 & 65.1 & 52.6 & 61.6 & 59.1 & 72.0 & 77.6 & 73.7 & 79.3 & 44.0   & 34.1 & 62.4 \\
			                                                        & medium    & 40.9 & 53.0 & 34.1 & 59.1 & 56.5 & 71.1 & 71.1 & 64.7 & 84.5 & 4.7    & 27.6 & 51.6 \\
			                                                        & coarse    & 65.1 & 53.9 & 44.4 & 52.2 & 66.4 & 64.7 & 69.8 & 72.0 & 75.9 & 6.9    & 3.4  & 52.2 \\
			\cline{2-14}
			\multirow[t]{3}{*}{zephyr-7b~\cite{tunstall2023zephyr}} & fine      & 67.7 & 43.1 & 60.3 & 60.3 & 56.5 & 83.2 & 78.4 & 79.3 & 86.2 & 62.5   & 28.0 & 64.1 \\
			                                                        & medium    & 66.4 & 40.5 & 82.8 & 81.5 & 48.7 & 85.3 & 46.1 & 71.6 & 85.3 & 3.9    & 12.9 & 56.8 \\
			                                                        & coarse    & 65.9 & 41.8 & 78.9 & 56.9 & 42.7 & 85.8 & 49.1 & 71.6 & 85.8 & 3.4    & 3.4  & 53.2 \\
			\bottomrule
		\end{tabular}
	}
\end{table*}

\subsubsection{RQ2: Denser Knowledge Improves Cross-Modality Performance}
Building upon the observation that knowledge injection significantly impacts model performance (RQ1), we investigate the influence of knowledge density. We hypothesize that richer medical information within captions will lead to enhanced cross-modality learning and improved classification accuracy. To test this, we leverage the three knowledge granularity levels defined in Section \ref{sec:sec:knowledge_injection_framework}: coarse-grained (disease labels only), medium-grained (disease labels with typical phenotypes), and fine-grained (disease labels with typical and excluded phenotypes). These levels represent increasing densities of medical knowledge embedded in the captions.

We fine-tuned CLIP on captions generated at each granularity level using three 7B LLMs (llama2, vicuna, and zephyr). Tab.~\ref{tab:differrnt_knowledge_density} shows the zero-shot classification accuracy on CheXpert. We observe a clear positive correlation between knowledge density and accuracy. For example, incorporating disease phenotypes (medium-grained) consistently yields an average 4.52\% accuracy improvement over using only disease labels (coarse-grained) ($p < 0.05$, paired t-test). Further, the inclusion of excluded phenotypes (fine-grained) results in a substantial 12.36\% average accuracy gain compared to the medium-grained level ($p < 0.01$, paired t-test). These findings support our hypothesis that denser knowledge significantly benefits cross-modality learning.

While all three LLMs demonstrate this trend, Zephyr consistently achieves the highest accuracy across all levels, potentially due to its training data incorporating more medical knowledge. We further explore the impact of domain-specific LLMs in RQ3. Notably, the accuracy achieved with human captions is comparable to the coarse-grained level, highlighting that human captions often lack the detailed phenotypic information that significantly benefits the model. This highlights our proposed knowledge injection framework in providing the model with crucial information for optimal performance.

In conclusion, our findings underscore the vital role of knowledge density in the effectiveness of knowledge injection for medical cross-modality learning. Denser knowledge, encompassing both typical and excluded phenotypes, leads to substantially improved classification accuracy. This emphasizes the importance of developing methods for systematically injecting detailed, medical knowledge into pre-trained models to enhance their performance in medical image analysis.

\subsection{RQ3: Factors Influencing Knowledge Injection Effectiveness}

Building upon the significant impact of knowledge (RQ1) and the benefit of denser knowledge (RQ2), we further analyze factors influencing knowledge injection effectiveness. We explore the role of domain-specific LLMs and the potential of integrating vision-language models (vLLMs) for caption generation, aligning with our goal of investigating the influence of knowledge sources on pre-trained model performance. 

\begin{table*}[t]
	\centering
	\caption{Zero-shot classification accuracy for different LLMs and vLLM integration strategies.} 
	\label{tab:domain_specific_llms_and_vllm}
	\resizebox{0.7\linewidth}{!}{
		\begin{tabular}{llrrrrrrrrrrrr}
			\toprule
			                                                              & diagnosis & A    & CM   & C    & E    & ECM  & LL   & LO   & PE   & PO                                 & P    & PTX  & Avg  \\
			Model                                                         & Grained   &      &      &      &      &      &      &      &      &                                    &      &      &      \\
			\midrule
			\multirow[t]{5}{*}{llama2-70b~\cite{touvron2023llama2}}       & fine      & 72.0 & 55.2 & 47.8 & 72.8 & 56.5 & 73.3 & 78.4 & 71.1 & 89.2                               & 3.9  & 31.5 & 59.2 \\
			                                                              & medium    & 62.1 & 54.3 & 46.1 & 65.5 & 56.5 & 67.7 & 73.7 & 78.0 & 75.9                               & 8.2  & 43.1 & 57.4 \\
			                                                              & coarse    & 62.1 & 54.3 & 41.4 & 44.8 & 65.1 & 72.0 & 67.2 & 65.1 & 79.3                               & 16.8 & 13.4 & 52.9 \\
			\cline{2-14}
			\multirow[t]{3}{*}{med42-70b~\cite{med42}}                    & fine      & 73.3 & 68.1 & 57.3 & 69.4 & 62.5 & 61.2 & 78.9 & 73.3 & 66.4                               & 33.6 & 40.1 & 62.2 \\
			                                                              & medium    & 59.5 & 50.4 & 44.8 & 70.7 & 56.5 & 72.8 & 76.7 & 74.6 & 84.5                               & 11.2 & 39.7 & 58.3 \\
			                                                              & coarse    & 65.9 & 29.3 & 70.7 & 19.4 & 43.5 & 81.5 & 57.3 & 70.7 & 81.0                               & 4.7  & 3.4  & 47.9 \\
			\cline{2-14}
			\multirow[t]{3}{*}{mistral-7b-01~\cite{jiang2023mistral}}     & fine      & 57.3 & 44.0 & 35.8 & 52.2 & 49.1 & 75.4 & 75.4 & 61.2 & 83.2                               & 11.6 & 27.6 & 52.1 \\
			                                                              & medium    & 51.3 & 55.2 & 44.0 & 55.6 & 65.5 & 57.8 & 69.8 & 66.4 & 70.3                               & 3.4  & 12.9 & 50.2 \\
			                                                              & coarse    & 65.9 & 29.7 & 81.5 & 69.8 & 44.0 & 85.8 & 45.3 & 71.6 & 85.8                               & 3.4  & 6.9  & 53.6 \\
			\cline{2-14}
			\multirow[t]{3}{*}{BioMistral-7B~\cite{labrak2024biomistral}} & fine      & 67.7 & 59.1 & 55.2 & 61.6 & 60.8 & 74.6 & 77.2 & 74.1 & 82.8                               & 47.4 & 50.9 & 64.7 \\
			                                                              & medium    & 56.9 & 56.5 & 41.8 & 59.5 & 60.3 & 59.1 & 71.6 & 72.4 & 75.9                               & 4.3  & 17.7 & 52.4 \\
			                                                              & coarse    & 36.6 & 53.9 & 37.1 & 58.2 & 59.1 & 65.1 & 69.4 & 61.6 & 72.0                               & 6.5  & 19.4 & 49.0 \\
			\midrule
			gemini-pro-vision~\cite{geminiteam2023gemini}                 & -         & 57.3 & 29.7 & 80.6 & 64.2 & 47.8 & 86.6 & 49.6 & 70.7 & 86.6                               & 28.0 & 19.8 & 56.4 \\
			chexagent~\cite{chen2024chexagent}                            & -         & 44.4 & 37.1 & 79.7 & 66.4 & 54.3 & 72.8 & 53.0 & 69.8 & 69.8                               & 39.2 & 35.8 & 56.6 \\
			\midrule
			\multirow[t]{3}{*}{BioMistral-7B-Gemini}                      & fine      & 42.2 & 60.3 & 48.7 & 64.2 & 60.8 & 52.6 & 67.7 & 66.8 & 66.4                               & 22.4 & 48.7 & 54.6 \\
			                                                              & medium    & 37.5 & 58.2 & 50.9 & 66.4 & 57.8 & 55.2 & 69.8 & 67.7 & 68.1                               & 19.8 & 53.0 & 54.9 \\
			                                                              & coarse    & 43.5 & 55.2 & 47.4 & 57.8 & 56.5 & 43.1 & 72.0 & 66.4 & 63.4                               & 5.6  & 39.7 & 50.1 \\
			\cline{2-14}
			\multirow[t]{3}{*}{BioMistral-7B-chexagent}                   & fine      & 64.2 & 62.1 & 60.3 & 59.9 & 74.1 & 36.6 & 79.3 & 73.7 & 39.2                               & 50.0 & 45.7 & 58.6 \\
			                                                              & medium    & 68.5 & 64.7 & 63.4 & 61.2 & 73.7 & 46.6 & 78.0 & 74.1 & 52.6                               & 65.1 & 59.5 & 64.3 \\
			                                                              & coarse    & 67.2 & 58.2 & 59.1 & 68.5 & 70.7 & 32.8 & 78.9 & 71.6 & 54.3                               & 40.5 & 55.2 & 59.7 \\
			\bottomrule
		\end{tabular}
	}
\end{table*}

\subsubsection{Impact of Domain-Specific LLMs}

Domain-specific LLMs, trained on extensive medical corpora, are hypothesized to possess a deeper understanding of medical concepts compared to general-purpose LLMs. This expertise could translate to more accurate and informative CXR captions, potentially enhancing cross-modality learning. To assess this, we compared two medical domain-specific LLMs, BioMistral~\cite{labrak2024biomistral} and Med42~\cite{med42}, against their base models, Mistral~\cite{jiang2023mistral} and Llama2~\cite{touvron2023llama2}, respectively.

Tab.~\ref{tab:domain_specific_llms_and_vllm} shows that BioMistral and Med42 generally achieve higher zero-shot classification accuracy on CheXpert compared to their base models, particularly with fine-grained knowledge injection. For example, BioMistral with fine-grained knowledge demonstrates a 7.77\% accuracy improvement over Mistral, highlighting the advantage of domain-specific LLMs in leveraging detailed medical knowledge for informative caption generation. However, this advantage diminishes with coarser knowledge levels, suggesting that domain expertise is most beneficial when handling complex information like fine-grained knowledge, which includes both typical and excluded phenotypes.

\subsubsection{Integrating vLLMs for Caption Generation}

vLLMs excel at generating descriptive image captions and enhancing cross-modal learning~\cite{liu2023visual,liu2023improved,chen2023sharegpt4v,dai2023instructblip}. However, their efficacy for medical cross-modality learning remains to be seen. We compared two vLLMs, Gemini-pro-vision \cite{geminiteam2023gemini} and ChexAgent \cite{chen2024chexagent}, against our knowledge injection approach using BioMistral. These vLLMs were chosen for their strong image captioning performance and relevance to medical imaging.

Tab.~\ref{tab:domain_specific_llms_and_vllm} reveals that while both Gemini-pro-vision and ChexAgent surpass the human caption baseline, they fall short of BioMistral with fine-grained knowledge injection. Furthermore, integrating vLLM-generated captions into our knowledge injection framework generally reduces accuracy compared to using BioMistral alone with the same knowledge level. This suggests that directly incorporating vLLM captions may introduce biases or noise that hinder the model's ability to effectively utilize the injected knowledge. For instance, vLLM captions might emphasize different image aspects compared to our knowledge, leading to informational inconsistencies. 

\section{Conclusion}
This study explored the impact of knowledge injection on medical cross-modality learning, focusing on CXR image classification. We introduced a novel Set Theory-based framework to generate captions with varying levels of medical knowledge granularity. Our findings demonstrate that incorporating fine-grained medical knowledge, particularly information about both typical and excluded phenotypes, substantially improves the performance of CLIP-based zero-shot classification on the CheXpert dataset. Notably, models fine-tuned on our generated captions significantly outperformed those trained using human-generated captions or pre-trained models lacking specific medical knowledge, highlighting the crucial role of domain expertise.

This study demonstrates the significant impact of injecting fine-grained, domain-specific knowledge on the performance of pre-trained models for medical cross-modality learning, particularly in CXR image classification. Our proposed framework, utilizing a set theory-based approach and domain-specific LLMs for caption generation, enables controllable knowledge injection and leads to substantial improvements in zero-shot classification accuracy. These findings highlight the importance of knowledge density and granularity, paving the way for developing more accurate and reliable diagnostic tools in medical imaging and other cross-modality applications. Future research can explore the generalizability of this framework to other medical imaging tasks and investigate alternative knowledge representation methods.

\bibliographystyle{IEEEtran}
\bibliography{refs}

\appendix
\section*{Preprint Notice}
© 2025 IEEE. Personal use of this material is permitted. Permission from IEEE must be obtained for all other uses, in any current or future media, including reprinting/republishing this material for advertising or promotional purposes, creating new collective works, for resale or redistribution to servers or lists, or reuse of any copyrighted component of this work in other works.

\end{document}